\documentclass[journal]{vgtc}                     

\onlineid{0}

\vgtccategory{Research}

\vgtcpapertype{please specify}

\title{Dynamic Worlds, Dynamic Humans: Generating Virtual Human-Scene Interaction Motion in Dynamic Scenes}

\author{%
  \authororcid{Yin Wang}{0009-0005-6088-3794},
  \authororcid{Zhiying Leng}{0000-0002-8773-6939}, 
  Haitian Liu, 
  \authororcid{Frederick W. B. Li}{0000-0002-4283-4228},
  Mu Li, 
  and 
  \authororcid{Xiaohui Liang}{0000-0001-6351-2538}
}

\authorfooter{
\item Yin Wang, Zhiying Leng, Haitian Liu and Mu Li are with Beihang University, Beijing, China. E-mail: wang\_yin@buaa.edu.cn 
    \item Frederick W. B. Li is with University of Durham, U.K.
  \item Xiaohui Liang is with the State Key Laboratory of Virtual Reality Technology and Systems, Beihang University, Beijing, China, and also with the Zhongguancun Laboratory, Beijing, China. (\textit{Corresponding author})

}

\abstract{%
Scenes are continuously undergoing dynamic changes in the real world. However, existing human–scene interaction generation methods typically treat the scene as static, which deviates from reality.
Inspired by world models, we introduce Dyn-HSI, the first cognitive architecture for dynamic human-scene interaction, which endows virtual humans with three humanoid components. 
(1) Vision (human eyes): we equip the virtual human with a Dynamic Scene-Aware Navigation, which continuously perceives changes in the surrounding environment and adaptively predicts the next waypoint. 
(2) Memory (human brain): we equip the virtual human with a Hierarchical Experience Memory, which stores and updates experiential data accumulated during training. This allows the model to leverage prior knowledge during inference for context-aware motion priming, thereby enhancing both motion quality and generalization.
(3) Control (human body): we equip the virtual human with Human–Scene Interaction Diffusion Model, which generates high-fidelity interaction motions conditioned on multimodal inputs.
To evaluate performance in dynamic scenes, we extend the existing static human–scene interaction datasets to construct a dynamic benchmark, Dyn-Scenes. We conduct extensive qualitative and quantitative experiments to validate Dyn-HSI, showing that our method consistently outperforms existing approaches and generates high-quality human–scene interaction motions in both static and dynamic settings. 
}


\teaser{
  \centering
  \begin{overpic}[width=\linewidth]{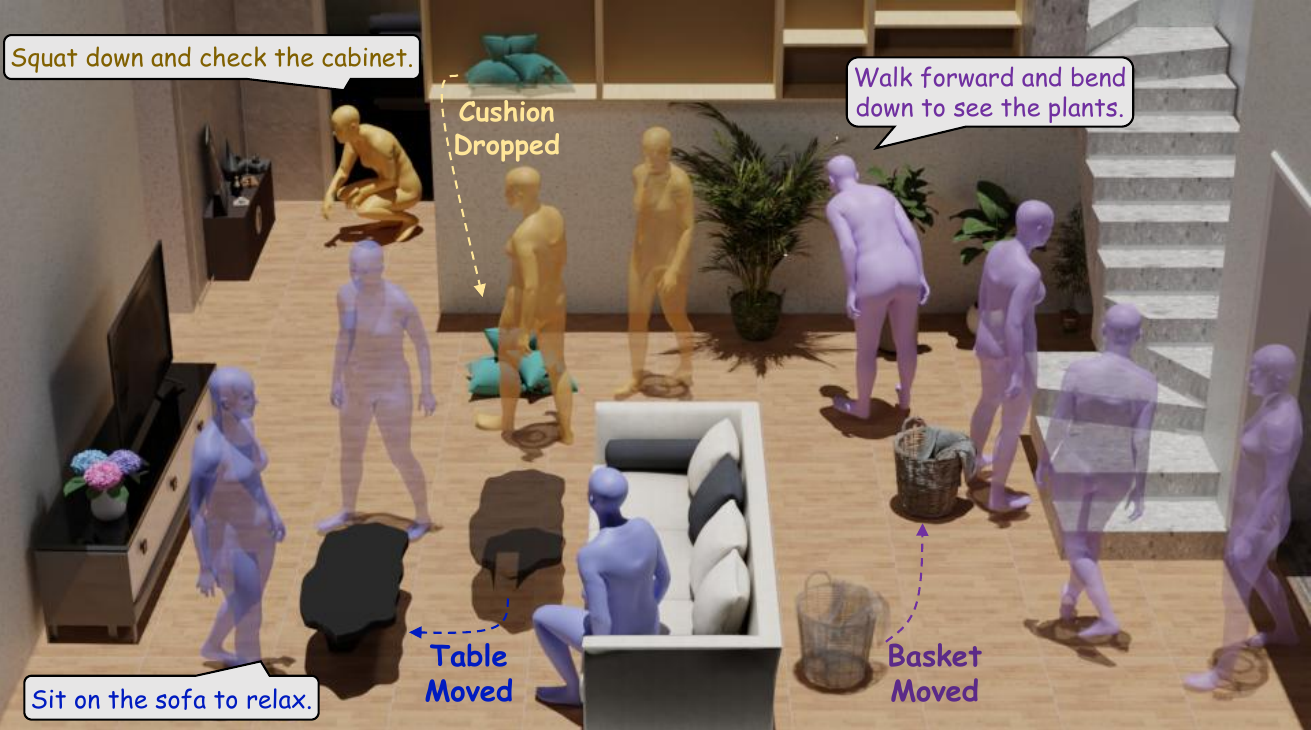}
  \end{overpic}
\caption{
Dyn-HSI is able to perceive scene changes and dynamically adjust virtual human motions to generate high-quality human–scene interactions. The arrows represent dynamic events occurring in the scene.
}
\label{fig:teaser}
\phantomsubcaption\label{teaser:a}
\phantomsubcaption\label{teaser:b}
\phantomsubcaption\label{teaser:c}
}

\graphicspath{{figs/}{figures/}{pictures/}{images/}{./}} 

\usepackage{times}                     

\usepackage{tabu}                      
\usepackage{booktabs}                  
\usepackage{lipsum}                    
\usepackage{mwe}                       
\usepackage{makecell}
\usepackage{pifont}
\usepackage{overpic}
\usepackage{color,colortbl}
\usepackage{amsmath} 
\usepackage{array,multirow,graphicx}

\usepackage{array}
\usepackage{arydshln}
\usepackage{tikz}
\usepackage{pict2e} 

\definecolor{skyblue}{RGB}{0,176,240}
\definecolor{brickred}{RGB}{237,125,49}
\usepackage{contour}

\usepackage{amssymb,bm}
\usepackage[ruled,linesnumbered,vlined]{algorithm2e}
\usepackage{graphicx}
\usepackage{cite}
\usepackage{xcolor}
\usepackage{tabularx}
\usepackage{multirow}
\usepackage{amsmath}
\usepackage{amssymb}
\usepackage{graphicx}
\usepackage{booktabs}
\usepackage[table]{xcolor}
\definecolor{lightgray}{gray}{0.9}
\definecolor{aliceblue}{rgb}{0.94, 0.97, 1.0}
\usepackage{arydshln}
\usepackage[normalem]{ulem}

\usepackage[font={sf,scriptsize}]{subcaption,caption}

\makeatletter
\renewcommand\paragraph{\@startsection{paragraph}{4}{1em}%
  {0ex \@plus 1ex \@minus.2ex}%
  {-1em}%
  {\reset@font\normalsize\sffamily\vgtc@sectionfont}}
\makeatother

\begin{document}


\firstsection{Introduction}

\maketitle

Driven by recent methodological advances \cite{bao2022rethinking,bao2025aucpro,bao2024improved}, virtual humans have made remarkable advances in content understanding \cite{jiang2024solami,jiang2023motiongpt}, appearance modeling \cite{liu2024humangaussian,hu2024gauhuman}, and motion generation \cite{bhattacharya2021speech2affectivegestures,bhattacharya2021text2gestures,liang2024intergen,li2025fine,zhong2023attt2m,wang2022humanise,wang2023fg,wang2025spatial,li2025x}, emerging as a crucial bridge between the physical and virtual worlds.
However, enabling virtual humans to interact effectively with their surrounding environments remains highly challenging—virtual humans should accurately navigate to target locations, interact with scene, and avoid physically implausible penetrations. While existing methods \cite{wang2022humanise,cen2024generating,wang2024move} have strived to improve human-scene interaction (HSI) quality, they rely on a simplifying assumption that the scene remains static, which can be restrictive in many real-world and dynamic interaction scenarios. In contrast, the physical world is inherently dynamic. As illustrated in Figure \ref{fig:teaser}, a cushion may fall to the ground or a table may be moved. This static assumption leads to a lack of perception and adaptability in dynamic scenes, resulting in unnatural penetration or failed interactions that severely diminish the sense of immersion and interactivity in virtual reality.

To address these challenges, we argue that virtual humans should be equipped with two additional humanoid core capabilities that go beyond existing methods. 
First, they need the ability to continuously perceive environmental dynamics, akin to human eyes. This enables them to acquire dynamic scene changes and avoid unreasonable collisions. Second, they require memory and generalization capabilities, akin to the human brain. This allows them to quickly adapt and generate reasonable interaction motions by recalling prior experiences, even when faced with new but similar scenes. Interestingly, this conceptual framework aligns perfectly with the theory of world models \cite{ha2018recurrent}. A world model is typically composed of a vision module, a memory module, and a controller, which precisely correspond to our perception-memory-generation loop. 

Building on this insight, we propose Dyn-HSI, a world model-based cognitive architecture to human-scene interaction in dynamic scenes, offering a novel and promising paradigm for addressing the challenges of virtual human interaction with dynamic worlds. Previous diffusion models methods \cite{zhang2022motiondiffuse,wang2024move} primarily rely on single-step to generate a complete motion sequence, which is limited to fixed, static inputs and fail to adapt when the scene changes. To overcome these limitations, we introduce a conditional autoregressive diffusion model. By iteratively incorporating dynamic conditions inputs, our model can generate interaction motions of arbitrary length while continuously adapting to scene changes.

More specifically, our framework consists of the following three key modules. 
\textbf{(1) Vision:} We equip virtual humans with a dynamic scene-aware navigation that continuously perceives and updates environmental changes, while leveraging voxelized human-scene occupancy information to intelligently plan the next path.
\textbf{(2) Memory:} We maintain a hierarchical experience memory that stores and updates experiential data accumulated during training. This enables context-aware motion priming as a form of test-time augmentation (TTA) during inference, allowing the model to leverage prior knowledge for improved decision-making and generalization.
\textbf{(3) Controller:} We feed multimodal inputs (e.g., scene, trajectory, text, etc.) to the controller to generate the corresponding human motion. Crucially, recognizing that different tasks exhibit varying dependencies on these conditions, we develop a conditional adapter. It adaptively adjusts the importance weights of different conditions based on the task's attributes. For example, a walking task places more emphasis on trajectory information, whereas a sitting task prioritizes scene constraints to prevent penetration.

Therefore, existing methods are fundamentally limited in dynamic scenes, as they treat the environment as static and thus fail to adapt to changing scene conditions. By contrast, our Vision–Memory–Controller architecture directly addresses these gaps: the vision enables virtual humans to continuously perceive scene dynamics; the memory allows them to accumulate and recall prior knowledge; and the controller plans and generates high-fidelity human–scene interaction motions.

Dyn-HSI benefits from the novel design of the dynamic scene-aware navigation, which iteratively focuses on the human's local voxelized human–scene occupancy. When scene changes occur, it can perceive the corresponding variations in the local voxelized occupancy and promptly adjust the trajectory to avoid collisions. Consequently, even though currently no dedicated human–scene interaction datasets exist for dynamic scenes, our approach generalizes effectively to such scenarios by training solely on static dataset LINGO \cite{jiang2024autonomous}, without requiring any additional training on dynamic scenes.
Furthermore, we manually construct a dynamic evaluation benchmark, Dyn-Scenes, based on the LINGO \cite{jiang2024autonomous} and Trumans \cite{jiang2024scaling} datasets, and conduct systematic evaluations on it to assess model performance in dynamic scenarios.
We also introduce a comprehensive evaluation protocol and metrics covering motion quality evaluation, trajectory evaluation, static scene evaluation, dynamic scene evaluation, and out-of-distribution scene evaluation. Extensive qualitative and quantitative experiments demonstrate that, compared with existing methods, Dyn-HSI achieves superior performance in both static and dynamic settings, highlighting its robustness capability. 
Our contributions are as follows:

\begin{itemize}
\item To the best of our knowledge, we are the first to tackle the task of virtual human–scene interaction in dynamic scenes.

\item We propose Dyn-HSI, the first cognitive architecture that equips virtual humans with Vision, Memory, and Controller.

\item We construct the Dyn-Scenes evaluation benchmarks, along with a comprehensive evaluation protocol and metrics, to assess model performance in dynamic scenes. 

\item Extensive experiments demonstrate that Dyn-HSI consistently achieves superior performance over existing methods in both static and dynamic settings.
\end{itemize}

\begin{figure*}[t]
    \centering
    \includegraphics[width=\linewidth]{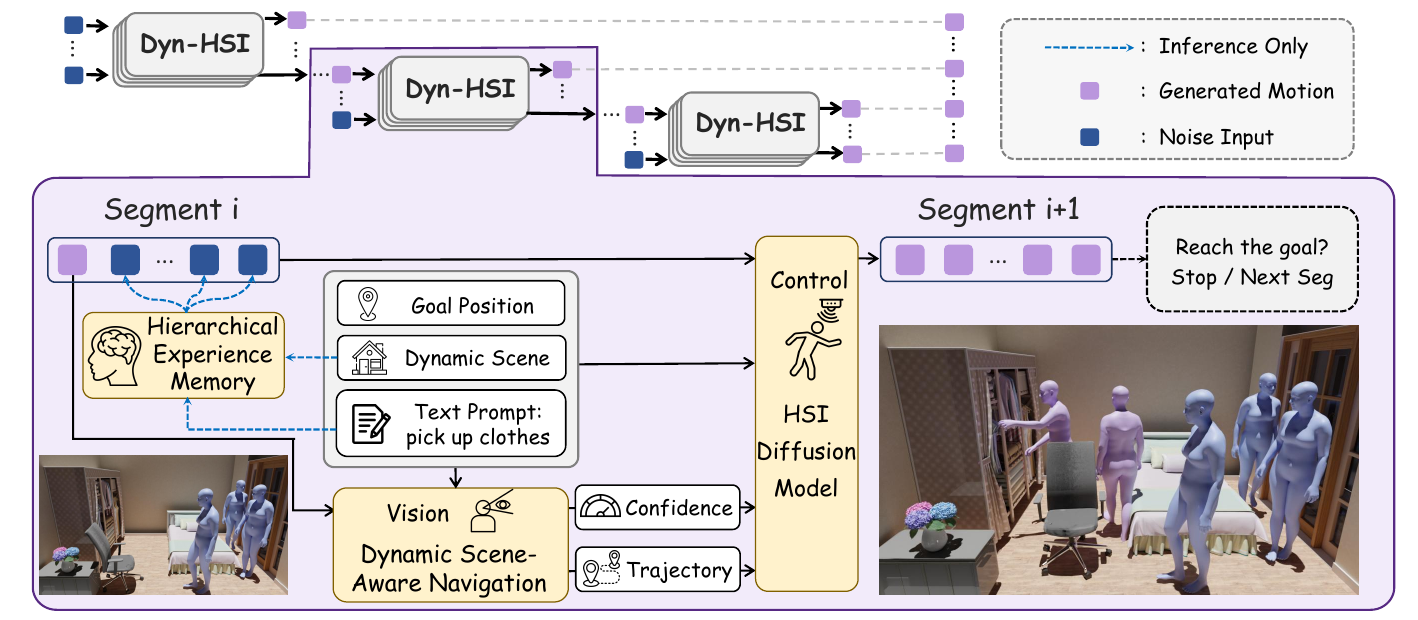}
    \caption{\textbf{Overview of Dyn-HSI.} Dyn-HSI is an autoregressive motion diffusion model capable of iteratively generating human–scene interaction motions. It consists of three components: (1) dynamic scene-aware navigation, (2) hierarchical experience memory, and (3) HSI diffusion model. }
    \label{pipeline}
\end{figure*}

\section{Related Work}
\subsection{Text-Driven Human Motion Generation} 
Three primary methodologies have emerged to tackle the challenge of text-driven single-human motion generation. (i) Latent space alignment \cite{ahuja2019language2pose,petrovich2022temos} aims to learn a unified latent space between textual and motion embeddings. (ii) Conditional autoregressive models \cite{zhang2023generating,guo2022tm2t,guo2023momask,pinyoanuntapong2023mmm} produce motion tokens in sequence, drawing on prior tokens and texts. (iii) Conditional diffusion models \cite{wang2025fg,chen2023executing,zhang2024large} have shown superior performance by leveraging the conditional diffusion framework to learn probabilistic text-motion mappings. 
Among them, some memory-based methods \cite{zhang2023remodiffuse,wang2025most} retrieve suitable motions from the training database to assist the motion generation process. However, such retrieval-based approaches require storing extremely large databases, which severely impacts inference time.
While these advancements have propelled human motion generation forward, they predominantly center on individual motion generation, lacking the ability to generate interactive motions with surrounding scenes.

\subsection{Text-Driven Human-Scene Interaction Generation}
Early work in this field mainly focused on single-frame human pose synthesis given a 3D scene configuration \cite{li2019putting,zhang2020place}. Zhao et al. \cite{zhao2022compositional} leveraged contact priors to generate human poses under text conditions. 
Xing et al. \cite{xing2024scene} proposed a scene-aware human motion prediction approach that constrains human poses using mutual distance representation to simulate human–scene interactions. However, prediction-based methods generally require several dozen past frames as historical priors and, in most cases, do not support text prompts, thereby creating a significant task gap compared to generation-based methods.
Building upon these studies, HUMANISE \cite{wang2022humanise} was the first to explore motion generation interacting with the scene under text conditions, combining human motion with scene point clouds and employing models such as PointNet \cite{qi2017pointnet++} to encode spatial constraints. Cen et al. \cite{cen2024generating} improved upon this by utilizing large language models (LLMs) to localize the target object's center coordinates in order to guide human motion generation. SceneDiffuser\cite{huang2023diffusion} leverages the iterative denoising process of diffusion models to integrate scene-aware generation, physics-based optimization into a unified framework. Sitcom-Crafter\cite{chen2024sitcom} introduces a modular motion generation system that combines human–scene and human–human interactions, using enhancement modules to synchronize motions and avoid collisions, thereby producing high-quality character animations. AMDM \cite{wang2024move} first predicts affordance maps of the scene and then generates corresponding human motion based on these maps. Recent works, such as the TRUMAN \cite{jiang2024scaling} and LINGO \cite{jiang2024autonomous} datasets, introduced voxel-based scene representations to enhance scene-aware motion generation for higher-quality human-scene interactions. 
However, current methods are limited to static-scene conditions, which results in a lack of perception and adaptation in dynamic scenes, causing unnatural penetrations or failed interactions.


\subsection{World Models}
The architecture of world models integrates vision, memory, and control modules with the aim of mimicking human thought and decision-making processes, playing a crucial role in understanding and predicting environmental dynamics. Early explorations of world model \cite{ha2018recurrent} mainly focused on gaming scenarios, where models were proposed to learn the spatiotemporal dynamics of game environments. Subsequently, research in the Dreamer series \cite{hafner2019dream,hafner2023mastering} further demonstrated the effectiveness of world models across a wide variety of gaming tasks. MineWorld \cite{guo2025mineworld} introduced a real-time interactive world model based on Minecraft, powered by a vision–action autoregressive Transformer that takes paired game scenes and actions as input to generate subsequent scenes conditioned on the actions. In recent years, autonomous driving has emerged as a major frontier for world model applications, with numerous methods \cite{hu2023gaia,wang2024drivedreamer} proposed to evaluate their effectiveness in driving scenarios. 
However, in the motion generative field, there has been little attention paid to how world models can be utilized.

\begin{figure*}[t]
    \centering
    \includegraphics[width=\linewidth]{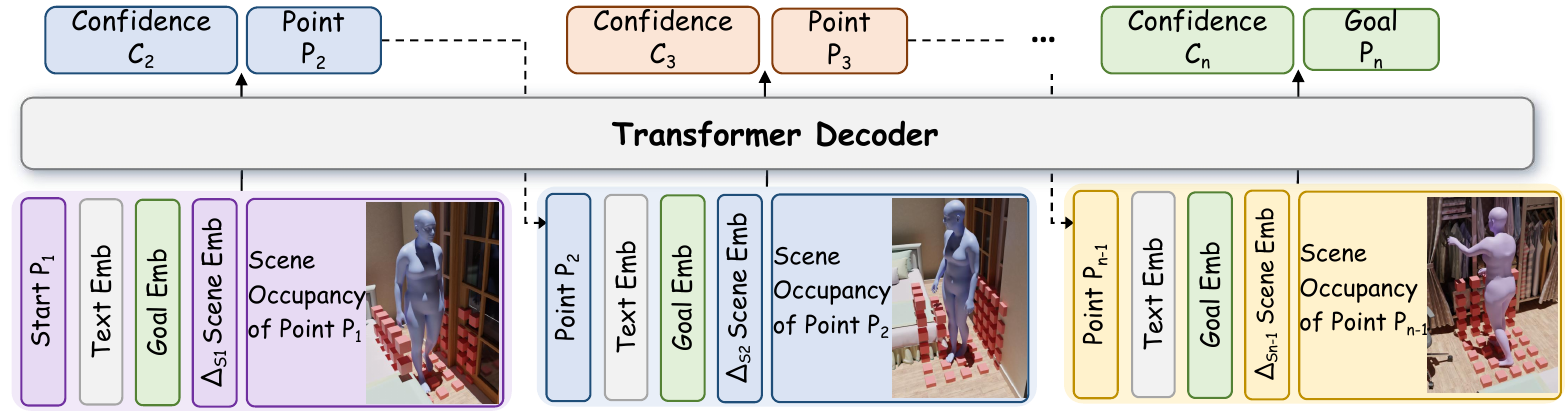}
    \vspace{-0.4cm}
    \caption{\textbf{Architecture of Dynamic Scene-Aware Navigation.} }
    \label{model1}
\end{figure*}

\section{Methods}

\subsection{Problem Formulation}
Formally, given a 3D scene $\mathrm{S}$, text prompt $\mathrm{T_f}$, and a goal position $\mathrm{G}$, our objective is to generate a human motion sequence $\mathrm{M}$ of length $L_m$ that interacts with the scene. 

For human motion, we adopt the SMPL-X \cite{pavlakos2019expressive} parametric model to represent the human motions $\mathrm{\{M_{i}\}_{i=1}^{L_m}}$ with joint positions form $\mathrm{M_{i}} \in \mathbb{R}^{J \times 3}$, where $J$ denotes the number of joints. These joints are fitted to the SMPL-X parametric model, which is then optimized to obtain the corresponding pose parameters $\theta_p$, global orientation $\theta_g$, hand poses $\theta_h$, and root translation $\theta_r$, thereby yielding the final posed human mesh.

For conditions, the 3D scene $\mathrm{S}$ at time $t$, denoted as $\mathrm{S}_t \in \{0,1\}^{L_s \times W_s \times H_s}$, is represented in a voxelized form, where 1 indicates that the position is occupied by the scene. Here, $L_s$, $W_s$, and $H_s$ denote the length, width, and height of the scene, respectively. Notably, $\mathrm{S}_t$ is dynamic and varies across different time $t$.
The text prompt is represented as $\mathrm{T_f} \in \mathbb{R}^{L_t}$, where $L_t$ is the dimension of the sentence vector. The goal position is represented as a 3D location $\mathrm{G} \in \mathbb{R}^{3}$ specified by the user.

\subsection{Overview and Technical Innovations}


As illustrated in Figure \ref{pipeline}, we introduce Dyn-HSI, an autoregressive motion diffusion model-based framework for text-driven dynamic human–scene interaction generation. 
In each segment, we start from a noisy human motion initialized close to Gaussian noise, represented as $X_{T}$. In particular, due to the autoregressive nature of the generation process, all generated human motions are conditioned on the previous two frames from the last round of human motion. Our Dyn-HSI framework receives the current 3D scene, the text prompt, and the segment's goal position as conditions to iteratively generate the clean human motion $X_{0}$ by reversing the diffusion process over $T$ steps. This procedure is repeated $T_N$ times, where $T_N$ denotes the total number of segments. All generated motions are concatenated to obtain the final complete human motion sequence.

Dyn-HSI is the first cognitive architecture for dynamic human–scene interaction. It can perceive scene changes and dynamically adjust virtual human–scene interaction motions through three interconnected modules working collaboratively: (1) The key innovation of the dynamic scene-aware navigation module is its ability to detect scene changes and adaptively predict the next waypoint, while also estimating a confidence score for each step to emphasize high-confidence trajectory points and reduce reliance on uncertain ones. (2) The key innovation of the hierarchical experience memory module lies in a hierarchical context-aware motion priming mechanism that is progressively organized from coarse actions to fine-grained actions and ultimately to scene-level contexts, enabling the retrieval of the most relevant noisy motion for optimal data initialization. (3) The key innovation of the HSI diffusion model is its ability to adaptively adjust the importance of each condition according to task attributes, thereby generating high-quality human–scene interaction motions under multimodal inputs.


\subsection{Vision: Dynamic Scene-Aware Navigation}
Existing methods typically treat the scene as a static input, without considering its dynamic nature, and thus face challenges in modeling dynamic human-scene interactions. To address this limitation, we aim to endow the virtual human with a form of ``vision" that enables it to dynamically perceive changes in the surrounding scene and make timely decisions. Thus, we introduce the Dynamic Scene-Aware Navigation, which dynamically predicts the next position based on the human's previous location.

Specifically, as illustrated in figure \ref{model1}, the scene occupancy input for each step is dynamically updated according to the human’s current position. At each decoding step, the model takes as input the current position, text embedding, goal embedding, scene variation between consecutive frames, and per-frame human-scene occupancy information, computed by aligning the human to the last coordinate point to capture the local occupancy relationship with the scene, in order to predict the next step in path planning.

For the current position $\mathrm{P}$, we first add positional embeddings to enable the model to capture the spatial context of the decoding step, and then encode it through an MLP to obtain the position feature $\mathrm{P_f} \in \mathbb{R}^{D_P}$, where $D_{P}$ represents the dimension of the position embeddings. For the text prompt, we employ the CLIP \cite{radford2021learning} to extract text features $\mathrm{T_p} \in \mathbb{R}^{D_T}$, where $D_{T}$ represents the dimension of the text embeddings. The goal position is encoded via an MLP to produce the goal embedding $\mathrm{G_f} \in \mathbb{R}^{D_G}$, where $D_{G}$ represents the dimension of the goal embeddings.


For scene occupancy, the entire scene is represented by a global occupancy grid $\mathrm{S_{global}}$, where each cell contains a boolean value indicating accessibility.
Since we only focus on predicting the next step, it is unnecessary to consider the entire scene.
Instead, we focus on the local occupancy around the human’s current position.
Inspired by LINGO \cite{jiang2024autonomous}, we construct a 3D voxel grid that encapsulates the local scene information around the human.
Accordingly, we construct a local occupancy grid $\mathrm{S_{local}}$ centered at the character’s pelvis and aligned with the character’s orientation at the current frame, with the human’s current position $(P_x, P_y, P_z)$ at the center.
Specifically, the local scene volume spans $[-0.6,\,0.6] \times [0.0,\,1.2] \times [-0.6,\,0.6]$ meters and is discretized into a $32 \times 32 \times 32$ voxel grid.
Occupancy values for each voxel are queried from the global scene grid to generate a binary 3D array for the local grid, where 1 indicates an occupied voxel and 0 indicates a free voxel.
We then employ a Vision Transformer architecture \cite{dosovitskiy2020image} to extract scene features $\mathrm{S_{f}} \in \mathbb{R}^{D_S}$ from this local occupancy data, where $D_{S}$ represents the dimension of the local scene embeddings.

For explicitly detect scene changes, we focus on the difference between the current local scene and the previous local scene, which is defined as: $\Delta_{S}^i = \mathrm{S_f^i} - \mathrm{S_f^{i-1}}$, where $i$ denotes the $i$-th frame and $\Delta_{S}$ represents the scene-change feature. 
If the positions at two consecutive frames remain unchanged, then $\Delta_{S}$ is zero. By explicitly modeling dynamic scene-change term, the model can more easily capture the notion of a dynamic scene, thereby facilitating dynamic scene-aware navigation.

Moreover, since the future scene at each frame is unknown, accurately predicting the exact positions of future frames is highly challenging. As a result, the model's predictions may be inherently uncertain. If the navigation strictly follows an uncertain predicted position, collisions or penetrations may occur. 
To alleviate this issue, we additionally use a prediction head to estimate a confidence score $\mathrm{C}$ for each step, which reflects the model's reliability in its current prediction. By allowing the HSI interaction diffusion model to assign greater weight to high-confidence trajectory points and reduce reliance on low-confidence ones, the model can better account for uncertainty and produce more reasonable navigation paths.

We employ the Transformer decoder to iteratively generate the next-step positions. For example, at the $i$-th frame, the position $\mathrm{P_{i+1}}$ and current confidence $\mathrm{C_{i+1}}$ are given by:
\begin{equation}
\mathrm{P_{i+1}}, \mathrm{C_{i+1}} =\text{Decoder}(\mathrm{P_{i}}, [\mathrm{T_f^i};\mathrm{G_f^i};\mathrm{S_f^i};\Delta_{S}^i]),
\end{equation}
where $[\cdot ; \cdot]$ indicates a concatenation of the input tensors.

To supervise the training of the Dynamic Scene-Aware Navigation, we minimize the l2 loss between the predicted trajectory and the ground-truth motion trajectory $\text{Traj}$. 
To supervise the training of confidence score, we define a soft target derived from the trajectory error: $\mathrm{C_{GT}} = \text{exp}(-\parallel \text{Traj} - \hat\epsilon_\theta\parallel_2))$, where $\hat\epsilon_\theta$ denotes the trajectory prediction. A lower trajectory error corresponds to a higher ground-truth confidence. The predicted confidence $\mathrm{C_{Pre}}$ is then optimized using a Binary Cross-Entropy loss. This is formulated as:
\begin{equation}
\mathcal{L}_\text{traj} = \mathbb{E} [\parallel  \text{Traj} - \hat\epsilon_\theta\parallel_2^2 ],\ \mathcal{L}_\text{conf} = \text{BCE}(\mathrm{C_{GT}} - {\mathrm{C_{Pre}}})
\end{equation}



\subsection{Memory: Hierarchical Experience Memory}
Establishing the concept of a ``brain'' enables virtual humans to rapidly adapt and generate plausible interactive motions by recalling past experiences, even when facing unseen scenes. This essentially corresponds to a Retrieval-Augmented Generation (RAG) task. In the motion generation domain, existing RAG-based approaches \cite{zhang2023remodiffuse,wang2025most} typically store raw motion data from the training set as a database, and retrieve motions that are textually similar as references to facilitate the generation process. However, we argue that this strategy is suboptimal for two main reasons. First, storing the entire training set is time-consuming and memory-intensive, which contradicts the requirement for fast adaptation in dynamic scenes. Second, the generated motion distribution tends to heavily bias toward the retrieved samples, which severely impedes motion diversity. Moreover, if retrieval fails, the motion generation process is likely to collapse as well. 

When overloaded with too much information, human memory often becomes blurry---we may recall a vague impression without clear details. Interestingly, we find this phenomenon closely resembles the forward process in diffusion models, where data progressively becomes noisier. Moreover, inspired by the findings \cite{zhou2024golden}, we observe that the choice of the initial noise significantly affects the quality of the generated results. Motivated by this observation, we design a hierarchical experience memory that stores noisy motion representations to enable context-aware motion priming, organized progressively from actions to scenes.
Specifically, it includes two main functions: storing memory and retrieving memory.

For storing memory, this process occurs only during training. We first clarify the data being stored: 
for a motion sequence $X_0$, we extract the verb from the text prompt as the storage key. Then we apply the forward diffusion process for $X_0$ to obtain its noisy motion $X_T$, and store $X_T$ along with the it's local scene and text prompt as the value.
At the end of each training iteration, the following factors are considered when deciding whether to store data: motion, scene, text prompt, and loss. Specifically, we set a loss threshold $\tau_{l}$. If the training loss exceeds $\tau_l$, it indicates that the model has not sufficiently learned the sample, and thus the data is discarded. If the loss is below $\tau_l$, we further evaluate the similarity between the current data and the stored memory using a weighted multi-modal similarity function:
\begin{equation}
\text{S}_{s} = \alpha_s \cdot \text{Sim}_{\text{scene}} + \beta_s \cdot \text{Sim}_{\text{joints}} + \gamma_s \cdot \text{Sim}_{\text{text}},
\end{equation}
where the $\text{Sim}$ function denotes the cosine similarity. $\text{Sim}_{\text{scene}}$ and $\text{Sim}_{\text{text}}$ represent the similarities computed from the corresponding feature embeddings extracted by the models, while $\text{Sim}_{\text{joints}}$ measures the similarity between the current motion sequence and the motions already stored in memory, $\alpha_s$, $\beta_s$, and $\gamma_s$ are hyperparameter. For a given text task, we store at most \textit{top-$k$} memories. When the memory is full, if the similarity of the current data is higher than the minimum similarity among stored samples, an update is performed by replacing the least similar sample with the new noisy data $\mathbf{x}_T$. This results in a hierarchical experience memory: the first layer captures coarse actions, such as walking or running; the second layer captures fine-grained variations of each action, such as walking slowly or quickly; and the third layer captures actions under different scene contexts, such as walking beside a table.

For retrieving memory, this process only occurs during inference. We perform context-aware motion priming based on the current starting scene. Specifically, we first match at the coarse action level by checking whether the task has been stored in memory via the key corresponding to the verb in the given text. If it has not been stored, a random Gaussian noise is returned. Next, we use the $\text{Sim}_{\text{text}}$ function to select the best fine-grained action variation according to the text prompt. Finally, the $\text{Sim}_{\text{scene}}$ function is used to retrieve the most suitable stored noisy motion data based on the current scene context.


\subsection{Control: Human-Scene Interaction Diffusion Model}

The Human-Scene Interaction Diffusion Model serves as the controller for the virtual human, whose primary responsibility is to generate plausible human motion that interacts appropriately with the scene given multi-modal inputs. However, effectively modeling this is non-trivial due to the large number of conditioning inputs. For example in this work, we consider four conditions: text, scene, goal, trajectory and its confidence score. 
We observe that different tasks exhibit different dependencies on these conditions. For example, walking tasks rely more heavily on trajectory information, whereas sitting tasks prioritize scene constraints to avoid penetration. Therefore, we develop a Condition Adapter within the Diffusion Model to adaptively adjust the importance of each condition according to the task attributes.

For condition adapter, the core goal is to assign different weights to the conditions based on the current task. Specifically, given a text feature $T_f$, we use an MLP to encode it into weight coefficients:
\begin{equation}
[\text{R}_{1},\text{R}_{2},\text{R}_{3},\text{R}_{4}] = \text{Softmax}(\text{MLP}(T_f)),
\end{equation}
where $\text{R}_{1},\text{R}_{2},\text{R}_{3},\text{R}_{4}$ correspond to the importance of scene, trajectory, text, and goal under the given task. Each condition is then multiplied by its corresponding coefficient to obtain the weighted condition feature. Additionally, each trajectory is further weighted by its confidence score to produce trajectory constraints with varying levels of influence.

\begin{table*}[t]
\centering
\caption{Comparisons to SoTA methods on the LINGO \cite{jiang2024autonomous} dataset. ``$\uparrow$'' denotes that higher is better. ``$\downarrow$'' denotes that lower is better.  ``w/o'' denotes ``without'' and indicates ablation variants.
We report the best and the second-best results in \textcolor{red}{Red} cells and \textcolor{blue}{blue} cells.}
\resizebox{\linewidth}{!}{%
\begin{tabular}{l c c c c c c c }
\toprule 
\multirow{2}{*}{Methods} &\multicolumn{3}{c}{Motion Evaluation} &\multicolumn{3}{c}{Trajectory Evaluation} &\multicolumn{1}{c}{Diversity Evaluation}\\
\cmidrule(l){2-4} \cmidrule(l){5-7} \cmidrule(l){8-8} 

&FID $\downarrow$ &MPJPE $\downarrow$&Foot Skating $\downarrow$ &Traj. sim. $\uparrow$&Traj. err. $\downarrow$& Goal. err. $\downarrow$ & Diversity $\uparrow$ \\

\midrule

MotionDiffuse~\cite{zhang2022motiondiffuse} 
& 0.363
& 0.114 
& 0.020  
& 55.36\% 
& 0.143
& 0.995 
& 4.208 \\

ReMoDiffuse~\cite{zhang2023remodiffuse} 
& 0.293
& 0.099 
& 0.029  
& 60.13\% 
& 0.139
& 1.022 
& 4.009 \\

Trumans~\cite{jiang2024scaling} 
& 0.282
& 0.093
& 0.031 
& 58.83\% 
& 0.129
& 0.863 
& 4.234 \\

LINGO~\cite{jiang2024autonomous} 
& 0.187
& 0.071 
& 0.028 
& 70.84\% 
& 0.091
& 0.743 
& 4.251 \\

\noalign{\vskip 3pt}
\hdashline
\noalign{\vskip 3pt}
Ours (Dyn-HSI)
& $\cellcolor{red!10} \textbf{0.092}$
& $\cellcolor{red!10} \textbf{0.040} $
& $\cellcolor{red!10} \textbf{0.016} $
& $\cellcolor{red!10} \textbf{90.01}\% $
& $\cellcolor{red!10} \textbf{0.021}$
& $\cellcolor{red!10} \textbf{0.027} $
& $\cellcolor{red!10} \textbf{4.649}$  \\

\rowcolor{gray!5} ~~~w/o Dynamic Scene-Aware Navigation
& $\cellcolor{blue!10} 0.147 $
& 0.060
& 0.019
& 77.60\%
& 0.055
& 0.153
&  $\cellcolor{blue!10}4.483$ \\
\rowcolor{gray!5} ~~~w/o Hierarchical Experience Memory
& 0.161
&  $\cellcolor{blue!10}0.056$
&  $\cellcolor{blue!10}0.017$
&  $\cellcolor{blue!10}85.69\%$
&  $\cellcolor{blue!10}0.030$
&  $\cellcolor{blue!10}0.049$
& 4.472 \\
\rowcolor{gray!5} ~~~w/o Condition Adapter
& 0.270
& 0.074
& 0.022 
& 73.83\%
& 0.038
& 0.064
& 4.384 \\

\bottomrule
\end{tabular}%
}

\label{main_table}
\end{table*}

\begin{table*}[t]
\centering
\caption{Static Scene Evaluation is conducted on the LINGO \cite{jiang2024autonomous} dataset. Dynamic Scene Evaluation is conducted on the Dyn-LINGO. ``$\downarrow$'' denotes that lower is better.  ``w/o'' denotes ``without'' and indicates ablation variants.
We report the best and the second-best results in \textcolor{red}{Red} cells and \textcolor{blue}{blue} cells.}
\resizebox{\linewidth}{!}{%
\begin{tabular}{l c c c c c c c c }
\toprule 
\multirow{2}{*}{Methods} &\multicolumn{4}{c}{Static Scene Evaluation} &\multicolumn{4}{c}{Dynamic Scene Evaluation}\\
\cmidrule(l){2-5} \cmidrule(l){6-9}

&Pene. Value $\downarrow$ &Pene. Rate $\downarrow$& Pene. Mean $\downarrow$ &Pene. Max $\downarrow$ &Pene. Value $\downarrow$ &Pene. Rate $\downarrow$& Pene. Mean $\downarrow$ &Pene. Max $\downarrow$ \\

\midrule

MotionDiffuse~\cite{zhang2022motiondiffuse} 
& 36.20
& 16.59\% 
& 1472 
& 3123
& 230.8
& 33.57\% 
& 3511 
& 6978\\

ReMoDiffuse~\cite{zhang2023remodiffuse} 
& 38.32
& 16.12\% 
& 1522 
& 3026
& 146.1
& 31.36\% 
& 3124 
& 6722\\

Trumans~\cite{jiang2024scaling} 
& 35.61
& 13.30\% 
& 1416 
& 3193 
& 98.85
& 20.59\% 
& 2156 
& 5347 \\

LINGO~\cite{jiang2024autonomous} 
& 34.20
& 13.37\%
& 1397
& 3083
& 48.84
& 17.39\%
& 1821
& 4963 \\

\noalign{\vskip 3pt}
\hdashline
\noalign{\vskip 3pt}
Ours (Dyn-HSI)
& $\cellcolor{red!10} \textbf{26.01}$
& $\cellcolor{red!10} \textbf{12.49}\%  $
& $\cellcolor{red!10} \textbf{1308} $
& $\cellcolor{red!10} \textbf{1534}$
& $\cellcolor{red!10} \textbf{39.19} $
& $\cellcolor{red!10} \textbf{15.77}\% $
& $\cellcolor{red!10} \textbf{1652}$
& $\cellcolor{red!10} \textbf{4525}$ \\

\rowcolor{gray!5} ~~~w/o Dynamic Scene-Aware Navigation
& 34.89
& 13.28\%
& 1485
& 1707
& 48.59
& 17.28\% 
& 1810
& 4891 \\
\rowcolor{gray!5} ~~~w/o Hierarchical Experience Memory
& 27.81
&  $\cellcolor{blue!10}12.97\%$
& 1389
&  $\cellcolor{blue!10}1591$
& 46.08
& $\cellcolor{blue!10} 16.33\%$
& $\cellcolor{blue!10} 1715$
& $\cellcolor{blue!10} 4751$\\
\rowcolor{gray!5} ~~~w/o Condition Adapter
&  $\cellcolor{blue!10}26.77$
& 13.85\%
&  $\cellcolor{blue!10}1322$
&  1605
& $\cellcolor{blue!10} 44.67$
& 16.78\% 
& 1758 
& 4995\\

\bottomrule
\end{tabular}%
}

\label{scenetable}
\end{table*}

We adopt a Transformer architecture as the backbone of the diffusion model. Multiple conditions are concatenated to form the conditioning input, which is then processed by the condition adapter and used as the \texttt{[CLS]} token of the Transformer. This guides the denoising process of the noisy motion data $X_t$. To supervise this motion generation process, we minimize the l2 loss between predicted and ground truth motions, denoted as 
\begin{equation}
\mathcal{L}_{motion}=\mathbb{E} [\parallel X_0 - \epsilon_\theta(\mathrm{X}_t, \mathrm{T_f},\mathrm{G_f},\mathrm{S_f},\mathrm{Traj}) \parallel_2^2]. 
\end{equation}
where $t$ represent timestep and $\epsilon_\theta(\mathrm{X}_t, \mathrm{T_f},\mathrm{G_f},\mathrm{S_f},\mathrm{Traj})$ denotes the model prediction. Therefore, the total training loss includes the $\mathcal{L}_{motion}$ loss, trajectory loss, and confidence loss: 
\begin{equation}
\mathcal{L} = \mathcal{L}_{motion} + \lambda_{t} \mathcal{L}_\text{traj} + \lambda_{c} \mathcal{L}_\text{conf}. 
\end{equation}
where $\lambda_{t}$ and $\lambda_{c}$ are the trade-off hyperparameters.

\subsection{Inference}
We provide a detailed description of the sampling and inference procedure of Dyn-HSI under an arbitrary dynamic scene. Given an initial scene $\mathrm{S}_1$, a text prompt, a start position, and a goal position, and assuming that the scene transitions to $\mathrm{S}_2$ at time $t_2$, the following procedure is applied. First, the A* algorithm is employed to project the 3D scene onto the X-Z plane by aggregating the occupancy along the Y-axis, resulting in a 2D occupancy grid. The grid is processed by normalization, dilation, and thresholding to account for the human's physical size. A suitable path in the current scene is then computed on this grid. 
While A* enables efficient shortest-path planning, it does not always produce the safest or most realistic navigation behavior, and exploring more advanced or optimized planners is left for future work.
Along this path, $k$ key points are uniformly selected to serve as the goal positions for each segment, where $k$ corresponds to the number of segments. Within each segment, noisy motion is first retrieved from the Hierarchical Experience Memory according to the current text task, and subsequently denoised. During the denoising process, the trajectory prediction model computes segment's trajectories based on the current position, goal position, local scene information, and text prompt. Notably, if the current time exceeds $t_2$, the local scene occupancy information is updated to reflect the changed scene $\mathrm{S}_2$. Finally, the diffusion model generates human motion within each segment conditioned on the goal positions, scene information, trajectory, and text prompt. The motions of all segments are then concatenated to form the complete motion sequence. This procedure enables the generation of high-quality human-scene interaction motions under dynamic scene conditions.

\section{Experiments}
\subsection{Datasets, Metrics and Implementation Details}

\paragraph{Datasets.} 
In the absence of dynamic human–scene interaction datasets, we first evaluate our method on the static benchmarks. To further assess its effectiveness in dynamic scenarios, we introduce a new dynamic human–scene interaction benchmark.

\textbf{Static HSI datasets.} We train and evaluate our method using the static scene dataset, LINGO \cite{jiang2024autonomous}, which provides rich human-scene interaction sequences. This dataset contains 16 hours of motion sequences captured across 120 unique indoor scenes, each accompanied by precise textual descriptions. Additionally, we conduct evaluations on the out-of-distribution static scene dataset, Trumans \cite{jiang2024scaling}, which records the motions of 7 participants performing various actions across 100 indoor scenes, resulting in 15 hours of high-quality motion capture data.

\begin{figure}[t]
    \centering
    \includegraphics[width=\linewidth]{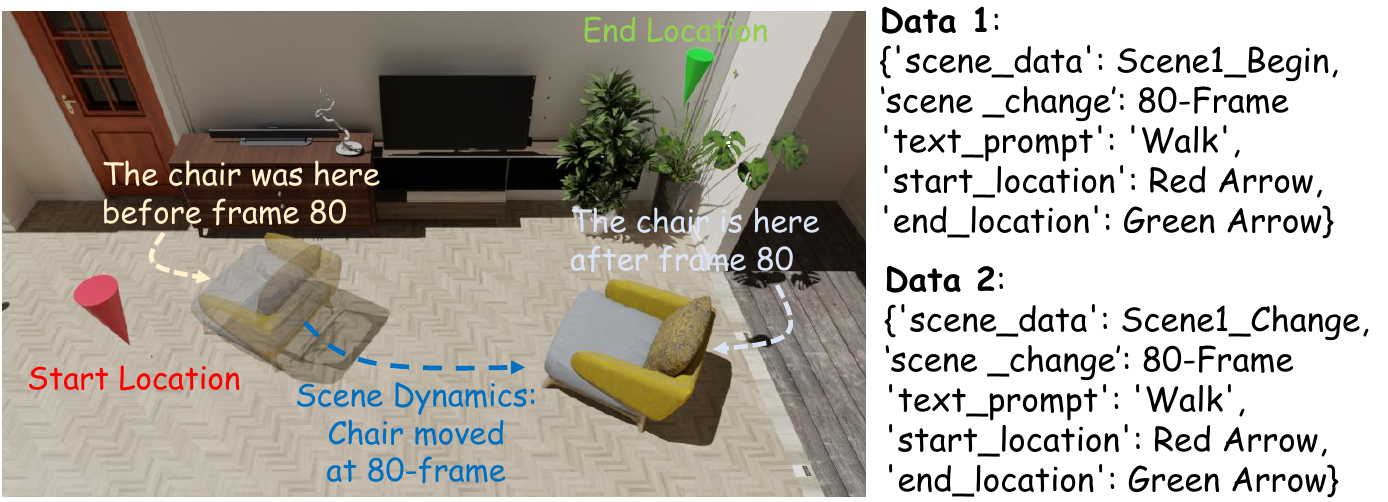}
    \caption{\textbf{Construction instructions for the dynamic scene evaluation benchmark.}  The recorded data is shown on the right.}
    \vspace{-0.2cm}
    \label{dataset}
\end{figure}

\textbf{Dynamic HSI benchmarks.} Notably, both datasets comprise only static scenes. To more rigorously evaluate the performance of Dyn-HSI under dynamic conditions, we manually modify the scenes in LINGO \cite{jiang2024autonomous} and Trumans \cite{jiang2024scaling} to generate updated scene occupancy data. 
Specifically, we annotate task compositions in dynamic scenes, which include the scene data, the time of scene changes, the text prompt, and the start and end locations. As illustrated in Figure \ref{dataset}, suppose the current scene is Scene 1. We define its initial state as Scene1\_Begin and export its voxel occupancy data. To do this, we first extract the scene’s point cloud, apply Poisson reconstruction to obtain a watertight mesh, and then use Kaolin’s check sign function \cite{KaolinLibrary} on voxelized query points to generate a voxel occupancy grid suitable for model input.
Next, we manually modify the scene, such as moving a chair, and export the updated voxel occupancy data, recording this as Scene1\_Change. The time of the change is marked at the 80-th frame. We also annotate the text prompt as a walking task and record the spatial coordinates of the red and green arrows as the starting and ending positions.
This procedure results in a dynamic-scene evaluation benchmark, consisting of approximately 70 modified scenes samples, referred to as Dyn-Scenes, which includes Dyn-LINGO and Dyn-Trumans.

\begin{table*}[t]
\centering
\caption{Out-of-Distribution Dynamic Scene Evaluation is conducted on the Dyn-Trumans benchmark. ``$\downarrow$'' denotes that lower is better.  ``w/o'' denotes ``without'' and indicates ablation variants.
We report the best and the second-best results in \textcolor{red}{Red} cells and \textcolor{blue}{blue} cells.}
\begin{tabular}{l c c c c c c}
\toprule 
\multirow{2}{*}{Methods} &\multicolumn{6}{c}{Out-of-Distribution Dynamic Scene Evaluation} 
\\
\cmidrule(l){2-7} 
&Foot Skating $\downarrow$ &Goal. err. $\downarrow$  &Pene. Value $\downarrow$ &Pene. Rate $\downarrow$& Pene. Mean $\downarrow$ &Pene. Max $\downarrow$\\

\midrule

MotionDiffuse~\cite{zhang2022motiondiffuse} 
& 0.020
& 4.227 
& 366.6 
& 45.50\%
& 4714 
& 7129\\

ReMoDiffuse~\cite{zhang2023remodiffuse} 
& 0.031
& 4.193 
& 326.4 
& 41.36\%
& 4691 
& 6734\\

Trumans~\cite{jiang2024scaling} 
& 0.033
& 3.532 
& 191.6 
& 29.46\% 
& 3085
& 6657
\\

LINGO~\cite{jiang2024autonomous} 
& $\cellcolor{blue!10}0.017$
& 1.138 
& $\cellcolor{blue!10}61.81$ 
& 18.53\% 
& 1723
& 6324 \\

\noalign{\vskip 3pt}
\hdashline
\noalign{\vskip 3pt}

Ours (Dyn-HSI)
& $\cellcolor{red!10} \textbf{0.010}$ 
& $\cellcolor{red!10} \textbf{0.946}$ 
& $\cellcolor{red!10}\textbf{41.44}$ 
& $\cellcolor{red!10} \textbf{16.05}\% $
& $\cellcolor{red!10} \textbf{1682}$ 
& $\cellcolor{red!10} \textbf{4849}$\\

\rowcolor{gray!5} ~~~w/o Dynamic Scene-Aware Navigation
& 0.042
& 1.201 
& 82.71 
& 18.52\% 
& 1940
& 6967
\\
\rowcolor{gray!5} ~~~w/o Hierarchical Experience Memory
& 0.045
& $\cellcolor{blue!10}0.972$ 
& 81.58 
& 17.22\% 
& 2013
& 7312 
\\
\rowcolor{gray!5} ~~~w/o Condition Adapter
& 0.033
& 1.018 
& 70.84 
& $\cellcolor{blue!10}16.98\%$ 
& $\cellcolor{blue!10}1710$
& $\cellcolor{blue!10}6162$
\\

\bottomrule
\end{tabular}%

\label{oodtab}
\end{table*}

\paragraph{Metrics.} 

We adopt several evaluation metrics from previous studies and further propose additional metrics.
First, we adopt the FID and Diversity metrics from \cite{guo2022generating}, which quantify the difference between the distributions of generated motions and ground-truth motions, and the dissimilarity among all generated motions across all textual descriptions, respectively. 
For a more comprehensive evaluation of human motion, we also report the mean per-joint position error (MPJPE) and adopt the foot skating metric proposed in \cite{he2022nemf} to assess foot sliding. 

To evaluate trajectory quality, we consider the goal position error (Goal. err.), trajectory error (Traj. err.), and trajectory similarity (Traj. sim.). The goal position error and trajectory error correspond to the average goal position error at the last frame and across all frames, respectively, which are defined as:
\begin{equation}
\mathrm{Goal. err.} = \big\lVert \mathbf{p}_T - \mathbf{p}^{*}_T \big\rVert_2, \ \mathrm{Traj. err.} 
= \frac{1}{T}\sum_{t=1}^{T} \big\lVert \mathbf{p}_t - \mathbf{p}^{*}_t \big\rVert_2 
\end{equation}
where $\mathbf{p}$ and $\mathbf{p}^{*}$ represent the predicted position and the ground-truth position, respectively. Trajectory similarity measures the fraction of frames for which the positional error is below a predefined threshold $\tau$. For each frame, the value is set to 1 if the error is less than $\tau$, and 0 otherwise. Formally, it is defined as:
\begin{equation}
\mathrm{Traj.\ sim.} 
= \frac{1}{T}\sum_{t=1}^{T} \mathbf{1}\big[\lVert \mathbf{p}_t - \mathbf{p}^{*}_t \rVert_2 < \tau\big],
\end{equation}

For human-scene interaction evaluation, we employ penetration-based metrics, including the penetration value (Pene. Value), penetration rate (Pene. Rate), average penetration (Pene. Mean), and maximum penetration (Pene. Max). 
The penetration value is defined as the mean squared per-frame fraction of mesh vertices intersecting the scene, emphasizing frames with larger collisions:
\begin{equation}
\mathrm{Pene. Value} 
=\lambda_v \frac{1}{T}\sum_{t=1}^{T} \left(\frac{1}{V_t}\sum_{i=1}^{V_t} \mathbb{I}_{t,i}\right)^{2} , 
\end{equation}
where $\lambda_v$ is a hyperparameter set to 100. $\mathbb{I}_{t,i}$ is the penetration indicator function, which equals 1 if the $i$-th vertex of the human mesh at frame $t$ intersects the scene and 0 otherwise, and $V_t$ denotes the number of human mesh vertices at frame $t$. The penetration rate measures the proportion of vertices intersecting the scene, averaged over all frames and vertices for each sample, which is defined as:
\begin{equation}
\mathrm{Pene. Rate} 
= \frac{\sum_{t=1}^{T}\sum_{i=1}^{V_t} \mathbb{I}_{t,i}}{\sum_{t=1}^{T} V_t},
\end{equation}

The average penetration quantifies the mean number of intersecting vertices per frame, while the maximum penetration indicates the largest number of intersecting vertices observed in any single frame, which are defined as:
\begin{equation}
\mathrm{Pene. Mean} 
= \frac{1}{T}\sum_{t=1}^{T}\sum_{i=1}^{V_t} \mathbb{I}_{t,i}, \ \mathrm{Pene. Max} 
= \max_{1 \le t \le T}\; \sum_{i=1}^{V_t} \mathbb{I}_{t,i} .
\end{equation}

\paragraph{Implementation Details.}
Regarding the Dynamic Scene-Aware Navigation, we adopt a 4-layer Transformer, where $D_P$, $D_T$, $D_G$, and $D_S$ are all set to 32. The size of $\mathrm{S_{local}}$ is set to $32 \times 32 \times 32$,  following the voxelization settings as LINGO~\cite{jiang2024autonomous} and Trumans~\cite{jiang2024scaling} to ensure consistency and fair comparison across methods.
For the Hierarchical Experience Memory, the hyperparameters $\alpha_s$, $\beta_s$, and $\gamma_s$ are set to 0.1, 0.4, and 0.5, $\alpha_r$ and $\gamma_r$ are set to 0.3 and 0.7, $\tau_{l}$ is set to 0.001, and $k$ is set to 200.
For the human-scene interaction diffusion model, we employ an 8-layer Transformer with a latent dimension of 512, generate 48 motion frames per segment, and set $\lambda_t$ to 0.5, $\lambda_c$ to 0.01.
In terms of the diffusion model, the variances $\beta_t$ are predefined to linearly spread from 0.0001 to 0.02, and the total number of noising steps is set at T = 100. We use the Adam optimizer with an learning rate of 0.0001. 
The training process is conducted on an Intel i9-10980XE CPU with 128,GB RAM and three NVIDIA GeForce RTX 3090 GPUs. The model is trained for 100 epochs, which takes approximately 2 days in total.
The batch size is set to 440 on a single GPU.


\subsection{Quantitative Results}

\paragraph{Evaluation of Human Motion Generation.}

The results in Table \ref{main_table} compare Dyn-HSI with state-of-the-art (SOTA) methods, including MotionDiffuse \cite{zhang2022motiondiffuse}, ReMoDiffuse \cite{zhang2023remodiffuse}, Trumans \cite{jiang2024scaling}, and LINGO \cite{jiang2024autonomous}, on the LINGO \cite{jiang2024autonomous} dataset. We evaluate human motion from three perspectives: (1) motion quality evaluation, which measures the fidelity of generated motions; (2) trajectory evaluation, which assesses the similarity between predicted and ground-truth (GT) trajectories; and (3) diversity evaluation, which quantifies the variety of generated motions. Overall, Dyn-HSI achieves consistently superior performance across all three aspects. Specifically, in terms of motion quality, Dyn-HSI generates motions that are closer to the real distribution, leading to better FID and MPJPE scores, surpassing Trumans by 67.45\% and 56.98\%, respectively, and further alleviating foot-sliding artifacts. For trajectory evaluation, Dyn-HSI achieves 90\% similarity with GT trajectories, indicating that our method produces more realistic and coherent motion paths. Notably, in terms of goal error, our method reduces the error by 96.36\% compared to LINGO \cite{jiang2024autonomous}, demonstrating that Dyn-HSI more accurately reaches the user-specified target positions. Finally, regarding diversity evaluation, Dyn-HSI achieves strong diversity scores, revealing that it generates high-quality motions without collapsing into a restricted distribution. 

\paragraph{Evaluation of Human-Scene Interaction.}

In Table \ref{scenetable}, we conduct two types of human–scene interaction evaluations for Dyn-HSI: (1) static scene evaluation, which assesses HSI quality on the original LINGO \cite{jiang2024scaling} dataset, and (2) dynamic scene evaluation, which evaluates HSI performance on our manually constructed Dyn-LINGO benchmark with scene variations. More specifically, under static scenes, Dyn-HSI outperforms existing methods in terms of penetration rate, average penetration, and maximum penetration—even against methods such as LINGO \cite{jiang2024autonomous} and Trumans \cite{jiang2024scaling} that are designed for static scenarios. Notably, our maximum penetration is reduced by 50.24\% compared to the best-performing method (LINGO), highlighting the superiority of Dyn-HSI in static environments. In dynamic scenes, methods like Trumans exhibit significant degradation, with penetration value increasing by 177.6\% and average penetration by 52.25\%. In contrast, Dyn-HSI maintains robust performance, with only a 50.67\% increase in penetration value and a 26.29\% increase in average penetration. Moreover, our penetration rate remains nearly identical to that in static settings, demonstrating that Dyn-HSI effectively handles dynamic scene conditions and delivers superior HSI quality under scene variations.

\paragraph{Evaluation of Out-of-Distribution Scenes.}

Since real-world is inherently diverse, scene variations cannot be fully covered by the training distribution. To better assess model performance under realistic conditions, we further conduct evaluations on the out-of-distribution (OoD) Dyn-Trumans benchmark, as reported in Table \ref{oodtab}. The results show that when shifting to OoD scenes, existing methods degrade substantially. For instance, Trumans \cite{jiang2024scaling} exhibits a 93.82\% increase in penetration value, while LINGO \cite{jiang2024autonomous} suffers a 27.42\% increase in maximum penetration. In contrast, Dyn-HSI demonstrates remarkable robustness to unseen scenarios: the penetration value rises by only 5.74\%, and the penetration rate increases by just 0.28\%. Meanwhile, Dyn-HSI achieves lower goal error and reduced foot-sliding compared to these methods. These findings confirm that our approach maintains strong effectiveness and robustness even on OoD data.

\begin{figure}[t]
    \centering
    \includegraphics[width=\linewidth]{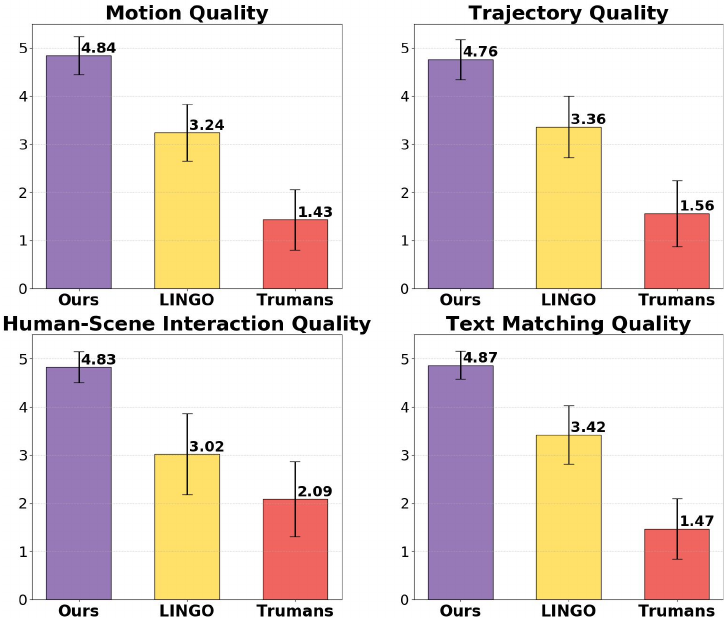}
    \caption{\textbf{User study results.} The higher score indicates better performance.}
    \label{userstudy}
\end{figure}

\begin{figure*}[t]
    \centering
    \includegraphics[width=\linewidth]{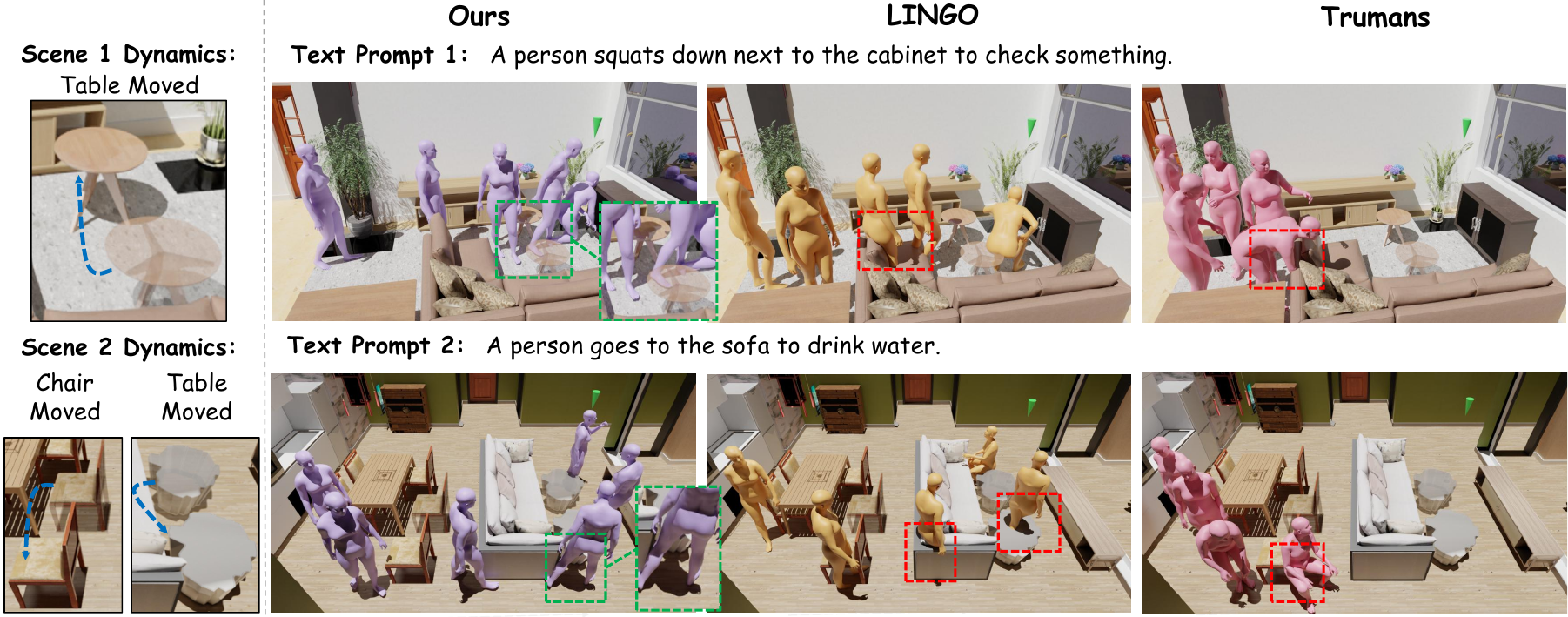}
    \caption{\textbf{Visual results compared with existing methods}. The left side provides a detailed description of the dynamic conditions within the scene. The green box zooms in on the details captured by our approach, while the red boxes highlight errors made by other methods. Green arrows indicate the goal positions.}
    \label{compare}
\end{figure*}

\subsection{User Study}

We also conducted a user study to more realistically evaluate the performance gap between methods, comparing Dyn-HSI with LINGO \cite{jiang2024autonomous} and Trumans \cite{jiang2024scaling}. The study involved 30 participants aged between 23 and 28, including 20 males and 10 females, among whom 15 were animation researchers. In this study, participants were asked to rate motions generated by the three methods, presented in randomized order. For each trial, participants viewed three human–scene interaction motions and rated them according to four criteria: (1) motion quality, (2) human–scene interaction quality, with a focus on penetration artifacts, (3) trajectory quality, with attention to unrealistic paths, and (4) consistency with the input text prompt. Each aspect was rated on a 1–5 scale with arbitrary values, and then the average score was calculated for final score. This study was approved by the authors' institutional ethical review board. Human subject testing was conducted at Beihang University and reviewed by the Beihang University Institutional Review Board for Biological and Medical Ethics (Protocol No. BM20250294).

\begin{figure}[t]
    \centering
    \includegraphics[width=\linewidth]{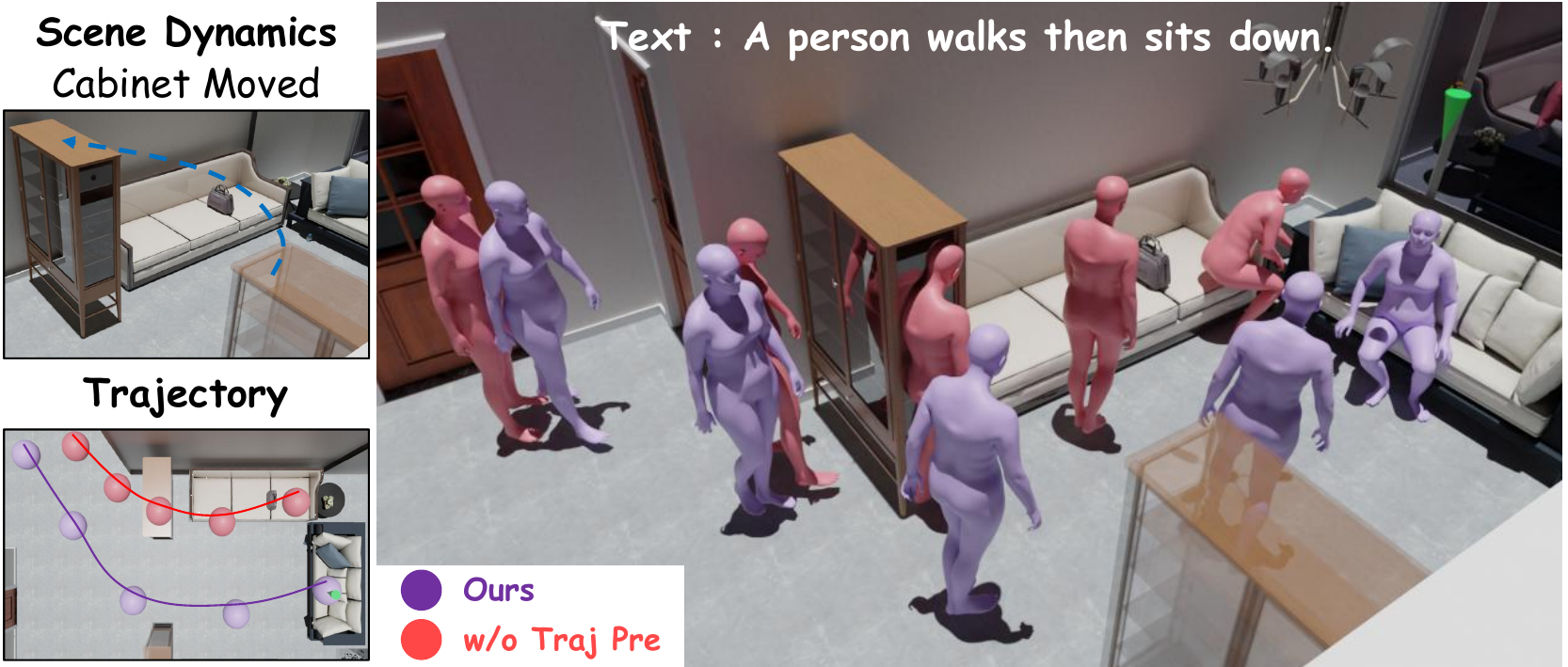}
    \caption{Qualitative examples from the ablation study on the Dynamic Scene-Aware Navigation. The bottom-left shows the top-down trajectory paths of both methods.}
    \label{study1}
    \vspace{-0.2cm}
\end{figure}

As shown in Figure \ref{userstudy}, Dyn-HSI consistently outperforms existing methods across all four criteria. For instance, in terms of text alignment, Dyn-HSI achieves an average score of 4.87, which is 42.39\% higher than LINGO \cite{jiang2024autonomous}, demonstrating that our method generates motions that more faithfully match textual descriptions and achieve superior semantic alignment. In addition, Dyn-HSI exhibits strong results in trajectory quality, motion quality, and HSI quality. 
To improve the methodological rigor of the subjective evaluation, we further conducted statistical analyses using the Friedman test and Kendall’s W on all four criteria, including motion quality ($\chi^2 = 15.5$, $p < 0.001$, $W = 0.94$), trajectory quality ($\chi^2 = 15.2$, $p < 0.001$, $W = 0.89$), human–scene interaction quality ($\chi^2 = 14.0$, $p < 0.001$, $W = 0.76$), and text-matching quality ($\chi^2 = 14.2$, $p < 0.001$, $W = 0.85$). These results indicate statistically significant differences among methods and strong inter-participant consistency.
These findings highlight that Dyn-HSI not only generates motions that closely follow textual prompts but also excels in modeling realistic human–scene interactions,
validating both the effectiveness and superiority of our approach.

\begin{figure}[t]
    \centering
    \includegraphics[width=\linewidth]{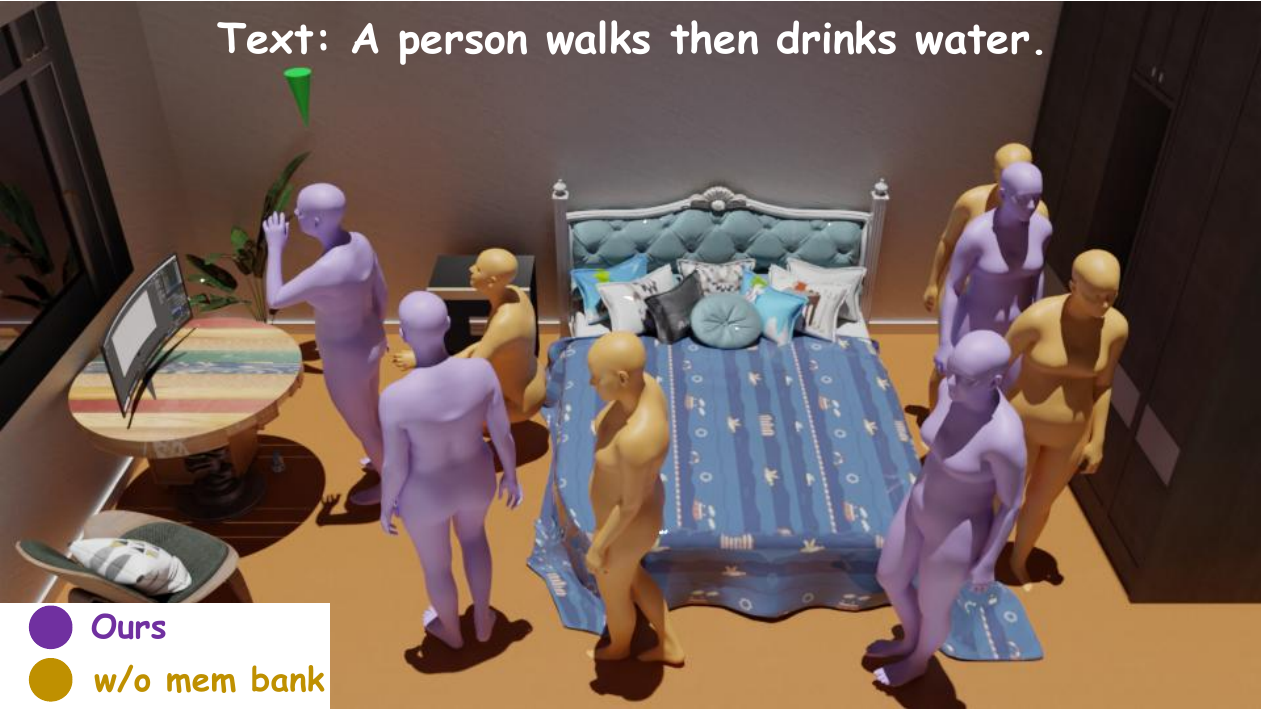}
    \caption{Qualitative examples from the ablation study on the Hierarchical Experience Memory.}
    \label{study2}
    \vspace{-0.2cm}
\end{figure}
\subsection{Qualitative Analysis}

In Figure \ref{compare}, we present qualitative comparisons of Dyn-HSI against LINGO \cite{jiang2024autonomous} and Trumans \cite{jiang2024scaling}. In the top example, the dynamic change in the scene is that the table is moved to the upper part of the environment. LINGO generates a motion that passes directly through the sofa, exhibiting severe penetration artifacts. Trumans fails under this setting, as it neither produces a reasonable motion nor reaches the specified location. In the bottom example, the dynamic scene involves two changes: a chair is moved and a table is moved. Under this setting, LINGO degrades significantly as it completely ignores both moved objects, failing to perceive the scene changes. Moreover, it does not follow the specified ``drinking'' action. Similarly, Trumans neither executes the intended action nor reaches the designated position. In contrast, Dyn-HSI successfully generates motions that are both semantically aligned with the text and highly interactive with the dynamic scene in both cases. For instance, in the first example, Dyn-HSI produces a motion that twists the body to step over the table, while in the second example it walks rightward to avoid the chair and then sidesteps to bypass the repositioned table. These results demonstrate that Dyn-HSI achieves fine-grained modeling of motion in dynamic human–scene interactions, highlighting the superiority of our approach.

\subsection{Ablation Study}

\paragraph{The Effectiveness of Dynamic Scene-Aware Navigation.}

\begin{table*}[t]
\centering
\caption{Efficiency analysis compared with SoTA methods. ``$\uparrow$'' denotes that higher is better. ``$\downarrow$'' denotes that lower is better.  ``w/o'' denotes ``without'' and indicates ablation variants.
We report the best and the second-best results in \textcolor{red}{Red} cells and \textcolor{blue}{blue} cells.}
\resizebox{0.7\linewidth}{!}{%
\begin{tabular}{l c c c c}
\toprule 
Methods &AITPF $\downarrow$   &FID $\downarrow$  &Traj. sim. $\uparrow$ &Pene. Value $\downarrow$\\

\midrule

MotionDiffuse \cite{zhang2022motiondiffuse}
& $\cellcolor{red!10} 0.175$
& 0.363 
& 55.36\% 
& 36.20\\

ReMoDiffuse \cite{zhang2023remodiffuse}
& 1.218
& 0.293 
& 60.13\% 
& 38.32\\

Trumans \cite{jiang2024scaling}
& 0.193
& 0.282 
& 58.83\%  
& 35.61\\

LINGO \cite{jiang2024autonomous}
& 0.234
& 0.187 
& 70.84\%  
& 34.20\\

Ours
& 0.413
& $\cellcolor{red!10}0.092$ 
& $\cellcolor{red!10}90.01\%$  
& $\cellcolor{red!10}26.01$\\

w/o Dynamic Scene-Aware Navigation 
& $\cellcolor{blue!10} 0.178$
& $\cellcolor{blue!10}0.147$
& 77.60\%  
& 34.89\\

w/o Hierarchical Experience Memory 
& 0.392
& 0.161
& $\cellcolor{blue!10}85.69\%$  
& 27.81\\

w/o Condition Adapter 
& 0.404
& 0.270 
& 73.83\%  
& $\cellcolor{blue!10}26.77$\\

\bottomrule
\end{tabular}%
}
\label{time}
\end{table*}

We perform a quantitative analysis, with results reported in Tables \ref{main_table}, \ref{scenetable}, and \ref{oodtab}. We observe that removing this module leads to a significant decline in trajectory-related performance. For example, in Table \ref{main_table}, trajectory similarity drops by 12.41\%, while trajectory error and goal error increase by 0.034 and 0.126, respectively. We also note that the degradation in trajectory quality poses substantial challenges for fine-grained human–scene interactions. In static scenes (Table \ref{scenetable}), the penetration value increases by 12.88, and in dynamic scenes, it increases by 9.40. These findings confirm the crucial role of Dynamic Scene-Aware Navigation in controlling human motion trajectories and enabling high-quality human–scene interactions.

We further conduct a qualitative ablation study of this module, as shown in Figure \ref{study1}. The trajectory plot in the bottom left corner shows that removing this critical module severely impairs the model's ability to perceive scene changes. Even when the cabinet moves from the bottom right to the top left, the motion of the red character exhibits substantial penetration with the object. In contrast, the full Dyn-HSI model effectively perceives scene changes and generates more reasonable human trajectories and human–scene interaction motions.

\paragraph{The Effectiveness of Hierarchical Experience Memory.}

We perform a quantitative analysis of the Hierarchical Experience Memory module. As shown in the Table \ref{main_table}, removing the Hierarchical Experience Memory leads to a decline in motion evaluation performance, indicating that optimized noise selection has a significant impact on the quality of generated motions. Moreover, results in Table \ref{oodtab} reveal that the Hierarchical Experience Memory has a strong effect on out-of-distribution data. For instance, the penetration value worsens by 96.86\%, the average penetration increases by 331, and the maximum penetration rises by 2463. This demonstrates that leveraging a known and familiar noise distribution can substantially improve generalization to unseen scenes.

Furthermore, we analyze the noise data stored in the Hierarchical Experience Memory. First, we note that our Hierarchical Experience Memory occupies only approximately 200 MB, which is significantly smaller compared to other memory-based motion generation methods ReMoDiffuse \cite{zhang2023remodiffuse} that utilize the entire training set (around 5 GB). Second, the Hierarchical Experience Memory contains approximately 400 entries spanning about 30 different actions. We believe that these noise motion data play a positive role in initializing certain specific motions. To illustrate, we examine the motion corresponding to the text prompt ``a person walks then drinks water," as shown in Figure \ref{study2}. Since the training data for drinking actions is relatively scarce, Dyn-HSI without the Hierarchical Experience Memory fails to generate this motion. In contrast, the full Dyn-HSI can effectively recall these motions and reproduce them with high quality. These results demonstrate the effectiveness of the Hierarchical Experience Memory module.

\paragraph{The Effectiveness of Condition Adapter.}

We also analyze the Condition Adapter in the Human–Scene Interaction Diffusion Model. Removing the Condition Adapter significantly degrades motion quality because the model cannot effectively differentiate the importance of different conditions. For example, the FID score increases 0.178 and MPJPE increases 0.034. Furthermore, we observe that removing the Condition Adapter causes an even larger decline in motion quality on out-of-distribution data. This is because the model fails to identify the most critical conditions when encountering unseen scenarios, leading to reduced motion fidelity. These results validate the importance of the Condition Adapter for maintaining high-quality motion generation.

\subsection{Efficiency Analysis}

We conduct an efficiency analysis evaluation, using the average inference time per frame (AITPF) as the primary metric. As shown in Table \ref{time}, MotionDiffuse \cite{zhang2022motiondiffuse} achieves the fastest inference speed, but its generation quality is notably poor, as it fails to effectively incorporate scene information. Trumans \cite{jiang2024scaling} and LINGO \cite{jiang2024autonomous} obtain moderate inference times. However, both methods neglect fine-grained interactions in dynamic scenes, resulting in unsatisfactory performance on trajectory similarity, penetration values, and other related metrics.
Our method is not the most time-efficient, yet it consistently achieves the best results in terms of motion quality and human–scene interaction. To further diagnose the internal bottlenecks, we analyze the runtime of our modules and observe that the Dynamic Scene-Aware Navigation is the major source of overhead. Since it iteratively predicts future waypoints, it introduces a significant time cost. When the Dynamic Scene-Aware Navigation is removed, the runtime becomes nearly comparable to MotionDiffuse. In contrast, the Hierarchical Experience Memory and condition adapter contribute negligible computational overhead.
Overall, our method requires only a small and acceptable increase in inference time, while delivering substantially higher accuracy and more realistic human–scene interaction modeling.



\section{Limitation and Future Work}

Here, we discuss the limitations of Dyn-HSI and suggest directions for future work.
First, although the generated motions can achieve fine-grained interactions with dynamic scenes, the inference speed is still not ideal. This limitation arises not only from the Dynamic Scene-Aware Navigation but also from the inherently slow inference of diffusion models.
Second, While the model generalizes to dynamic scenes, the current dynamic scenarios are manually simulated and do not fully capture the complexity of real-world dynamics. As a result, the system may encounter failure cases in highly complex or continuously changing environments, such as subtle object motion, crowded scenes, or long-term error accumulation.
Nevertheless, the construction of large-scale datasets that explicitly capture human–scene interactions in dynamic environments would greatly advance this line of research.
Extending the framework to more realistic and continuously dynamic settings introduces additional challenges, including reliable perception and tracking of moving objects or agents, more frequent scene updates, and tighter integration among perception, memory, and control modules to ensure stable long-term behavior.

\section{Conclusion}

In this paper, we address the challenging task of text-driven human–scene interaction motion generation under dynamic scene conditions. We propose Dyn-HSI, the first cognitive architecture for dynamic human-scene interaction that incorporates three key insights:
(1) Vision: we equip the virtual human with a Dynamic Scene-Aware Navigation, which continuously perceives changes in the surrounding environment and adaptively predicts the next waypoint;
(2) Memory: we introduce a Hierarchical Experience Memory, which optimizes the selection of initial noise by retrieving relevant noisy motion data; 
(3) Control: we design a HSI Diffusion Model, which assigns adaptive weights to different conditions, allowing the model to generate realistic interaction motions under multimodal inputs.
Extensive qualitative and quantitative experiments demonstrate that our method significantly outperforms existing approaches, achieving fine-grained human–scene interaction motion generation in both static and dynamic scenes.

\acknowledgments{%
This work was supported by the Academic Excellence Foundation of BUAA for PhD Students, the National Natural Science Foundation of China (Project Number: 62272019), the China Postdoctoral Science Foundation (Grant Number: 2025M774236), and the Postdoctoral Fellowship Program of CPSF (Grant Number: GZC20242159).
}

\bibliographystyle{abbrv-doi-hyperref}

\bibliography{template}

\begin{thebibliography}{10}

\bibitem{ahuja2019language2pose}
C.~Ahuja and L.-P. Morency.
\newblock Language2pose: Natural language grounded pose forecasting.
\newblock In {\em 2019 International Conference on 3D Vision (3DV)}, pp. 719--728. IEEE, 2019.

\bibitem{bao2022rethinking}
S.~Bao, Q.~Xu, Z.~Yang, X.~Cao, and Q.~Huang.
\newblock Rethinking collaborative metric learning: Toward an efficient alternative without negative sampling.
\newblock {\em IEEE Transactions on Pattern Analysis and Machine Intelligence}, 45(1):1017--1035, 2022.

\bibitem{bao2024improved}
S.~Bao, Q.~Xu, Z.~Yang, Y.~He, X.~Cao, and Q.~Huang.
\newblock Improved diversity-promoting collaborative metric learning for recommendation.
\newblock {\em IEEE Transactions on Pattern Analysis and Machine Intelligence}, 46(12):9004--9022, 2024.

\bibitem{bao2025aucpro}
S.~Bao, Q.~Xu, Z.~Yang, Y.~He, X.~Cao, and Q.~Huang.
\newblock Aucpro: Auc-oriented provable robustness learning.
\newblock {\em IEEE Transactions on Pattern Analysis and Machine Intelligence}, 2025.

\bibitem{bhattacharya2021speech2affectivegestures}
U.~Bhattacharya, E.~Childs, N.~Rewkowski, and D.~Manocha.
\newblock Speech2affectivegestures: Synthesizing co-speech gestures with generative adversarial affective expression learning.
\newblock In {\em Proceedings of the 29th ACM International Conference on Multimedia}, pp. 2027--2036, 2021.

\bibitem{bhattacharya2021text2gestures}
U.~Bhattacharya, N.~Rewkowski, A.~Banerjee, P.~Guhan, A.~Bera, and D.~Manocha.
\newblock Text2gestures: A transformer-based network for generating emotive body gestures for virtual agents.
\newblock In {\em 2021 IEEE virtual reality and 3D user interfaces (VR)}, pp. 1--10. IEEE, 2021.

\bibitem{cen2024generating}
Z.~Cen, H.~Pi, S.~Peng, Z.~Shen, M.~Yang, S.~Zhu, H.~Bao, and X.~Zhou.
\newblock Generating human motion in 3d scenes from text descriptions.
\newblock In {\em Proceedings of the IEEE/CVF conference on computer vision and pattern recognition}, pp. 1855--1866, 2024.

\bibitem{chen2024sitcom}
J.~Chen, P.~Hu, X.~Chang, Z.~Shi, M.~Kampffmeyer, and X.~Liang.
\newblock Sitcom-crafter: A plot-driven human motion generation system in 3d scenes.
\newblock {\em arXiv preprint arXiv:2410.10790}, 2024.

\bibitem{chen2023executing}
X.~Chen, B.~Jiang, W.~Liu, Z.~Huang, B.~Fu, T.~Chen, and G.~Yu.
\newblock Executing your commands via motion diffusion in latent space.
\newblock In {\em Proceedings of the IEEE/CVF Conference on Computer Vision and Pattern Recognition}, pp. 18000--18010, 2023.

\bibitem{dosovitskiy2020image}
A.~Dosovitskiy, L.~Beyer, A.~Kolesnikov, D.~Weissenborn, X.~Zhai, T.~Unterthiner, M.~Dehghani, M.~Minderer, G.~Heigold, S.~Gelly, et~al.
\newblock An image is worth 16x16 words: Transformers for image recognition at scale.
\newblock {\em arXiv preprint arXiv:2010.11929}, 2020.

\bibitem{KaolinLibrary}
C.~Fuji~Tsang, M.~Shugrina, J.~F. Lafleche, O.~Perel, C.~Loop, T.~Takikawa, V.~Modi, A.~Zook, J.~Wang, W.~Chen, T.~Shen, J.~Gao, K.~M. Jatavallabhula, E.~Smith, A.~Rozantsev, S.~Fidler, G.~State, J.~Gorski, T.~Xiang, J.~Li, M.~Li, and R.~Lebaredian.
\newblock Kaolin: A pytorch library for accelerating 3d deep learning research.

\bibitem{guo2023momask}
C.~Guo, Y.~Mu, M.~G. Javed, S.~Wang, and L.~Cheng.
\newblock Momask: Generative masked modeling of 3d human motions.
\newblock {\em arXiv preprint arXiv:2312.00063}, 2023.

\bibitem{guo2022generating}
C.~Guo, S.~Zou, X.~Zuo, S.~Wang, W.~Ji, X.~Li, and L.~Cheng.
\newblock Generating diverse and natural 3d human motions from text.
\newblock In {\em CVPR}, pp. 5152--5161, 2022.

\bibitem{guo2022tm2t}
C.~Guo, X.~Zuo, S.~Wang, and L.~Cheng.
\newblock Tm2t: Stochastic and tokenized modeling for the reciprocal generation of 3d human motions and texts.
\newblock In {\em European Conference on Computer Vision}, pp. 580--597. Springer, 2022.

\bibitem{guo2025mineworld}
J.~Guo, Y.~Ye, T.~He, H.~Wu, Y.~Jiang, T.~Pearce, and J.~Bian.
\newblock Mineworld: a real-time and open-source interactive world model on minecraft.
\newblock {\em arXiv preprint arXiv:2504.08388}, 2025.

\bibitem{ha2018recurrent}
D.~Ha and J.~Schmidhuber.
\newblock Recurrent world models facilitate policy evolution.
\newblock {\em Advances in neural information processing systems}, 31, 2018.

\bibitem{hafner2019dream}
D.~Hafner, T.~Lillicrap, J.~Ba, and M.~Norouzi.
\newblock Dream to control: Learning behaviors by latent imagination.
\newblock {\em arXiv preprint arXiv:1912.01603}, 2019.

\bibitem{hafner2023mastering}
D.~Hafner, J.~Pasukonis, J.~Ba, and T.~Lillicrap.
\newblock Mastering diverse domains through world models.
\newblock {\em arXiv preprint arXiv:2301.04104}, 2023.

\bibitem{he2022nemf}
C.~He, J.~Saito, J.~Zachary, H.~Rushmeier, and Y.~Zhou.
\newblock Nemf: Neural motion fields for kinematic animation.
\newblock {\em Advances in Neural Information Processing Systems}, 35:4244--4256, 2022.

\bibitem{hu2023gaia}
A.~Hu, L.~Russell, H.~Yeo, Z.~Murez, G.~Fedoseev, A.~Kendall, J.~Shotton, and G.~Corrado.
\newblock Gaia-1: A generative world model for autonomous driving.
\newblock {\em arXiv preprint arXiv:2309.17080}, 2023.

\bibitem{hu2024gauhuman}
S.~Hu, T.~Hu, and Z.~Liu.
\newblock Gauhuman: Articulated gaussian splatting from monocular human videos.
\newblock In {\em Proceedings of the IEEE/CVF Conference on Computer Vision and Pattern Recognition}, pp. 20418--20431, 2024.

\bibitem{huang2023diffusion}
S.~Huang, Z.~Wang, P.~Li, B.~Jia, T.~Liu, Y.~Zhu, W.~Liang, and S.-C. Zhu.
\newblock Diffusion-based generation, optimization, and planning in 3d scenes.
\newblock In {\em Proceedings of the IEEE/CVF Conference on Computer Vision and Pattern Recognition}, pp. 16750--16761, 2023.

\bibitem{jiang2023motiongpt}
B.~Jiang, X.~Chen, W.~Liu, J.~Yu, G.~Yu, and T.~Chen.
\newblock Motiongpt: Human motion as a foreign language.
\newblock {\em Advances in Neural Information Processing Systems}, 36:20067--20079, 2023.

\bibitem{jiang2024solami}
J.~Jiang, W.~Xiao, Z.~Lin, H.~Zhang, T.~Ren, Y.~Gao, Z.~Lin, Z.~Cai, L.~Yang, and Z.~Liu.
\newblock Solami: Social vision-language-action modeling for immersive interaction with 3d autonomous characters.
\newblock {\em arXiv preprint arXiv:2412.00174}, 2024.

\bibitem{jiang2024autonomous}
N.~Jiang, Z.~He, Z.~Wang, H.~Li, Y.~Chen, S.~Huang, and Y.~Zhu.
\newblock Autonomous character-scene interaction synthesis from text instruction.
\newblock In {\em SIGGRAPH Asia 2024 Conference Papers}, pp. 1--11, 2024.

\bibitem{jiang2024scaling}
N.~Jiang, Z.~Zhang, H.~Li, X.~Ma, Z.~Wang, Y.~Chen, T.~Liu, Y.~Zhu, and S.~Huang.
\newblock Scaling up dynamic human-scene interaction modeling.
\newblock In {\em Proceedings of the IEEE/CVF Conference on Computer Vision and Pattern Recognition}, pp. 1737--1747, 2024.

\bibitem{li2025x}
H.~Li, Z.~Wang, W.~Liang, and Y.~Wang.
\newblock X's day: Personality-driven virtual human behavior generation.
\newblock {\em IEEE Transactions on Visualization and Computer Graphics}, 2025.

\bibitem{li2025fine}
M.~Li, Y.~Wang, Z.~Leng, J.~Liu, F.~W. Li, and X.~Liang.
\newblock Fine-grained text-driven dual-human motion generation via dynamic hierarchical interaction.
\newblock {\em arXiv preprint arXiv:2510.08260}, 2025.

\bibitem{li2019putting}
X.~Li, S.~Liu, K.~Kim, X.~Wang, M.-H. Yang, and J.~Kautz.
\newblock Putting humans in a scene: Learning affordance in 3d indoor environments.
\newblock In {\em Proceedings of the IEEE/CVF conference on computer vision and pattern recognition}, pp. 12368--12376, 2019.

\bibitem{liang2024intergen}
H.~Liang, W.~Zhang, W.~Li, J.~Yu, and L.~Xu.
\newblock Intergen: Diffusion-based multi-human motion generation under complex interactions.
\newblock {\em International Journal of Computer Vision}, pp. 1--21, 2024.

\bibitem{liu2024humangaussian}
X.~Liu, X.~Zhan, J.~Tang, Y.~Shan, G.~Zeng, D.~Lin, X.~Liu, and Z.~Liu.
\newblock Humangaussian: Text-driven 3d human generation with gaussian splatting.
\newblock In {\em Proceedings of the IEEE/CVF Conference on Computer Vision and Pattern Recognition}, pp. 6646--6657, 2024.

\bibitem{pavlakos2019expressive}
G.~Pavlakos, V.~Choutas, N.~Ghorbani, T.~Bolkart, A.~A. Osman, D.~Tzionas, and M.~J. Black.
\newblock Expressive body capture: 3d hands, face, and body from a single image.
\newblock In {\em Proceedings of the IEEE/CVF conference on computer vision and pattern recognition}, pp. 10975--10985, 2019.

\bibitem{petrovich2022temos}
M.~Petrovich, M.~J. Black, and G.~Varol.
\newblock Temos: Generating diverse human motions from textual descriptions.
\newblock In {\em ECCV}, pp. 480--497, 2022.

\bibitem{pinyoanuntapong2023mmm}
E.~Pinyoanuntapong, P.~Wang, M.~Lee, and C.~Chen.
\newblock Mmm: Generative masked motion model.
\newblock {\em arXiv preprint arXiv:2312.03596}, 2023.

\bibitem{qi2017pointnet++}
C.~R. Qi, L.~Yi, H.~Su, and L.~J. Guibas.
\newblock Pointnet++: Deep hierarchical feature learning on point sets in a metric space.
\newblock {\em Advances in neural information processing systems}, 30, 2017.

\bibitem{radford2021learning}
A.~Radford, J.~W. Kim, C.~Hallacy, A.~Ramesh, G.~Goh, S.~Agarwal, G.~Sastry, A.~Askell, P.~Mishkin, J.~Clark, et~al.
\newblock Learning transferable visual models from natural language supervision.
\newblock In {\em International conference on machine learning}, pp. 8748--8763. PMLR, 2021.

\bibitem{wang2024drivedreamer}
X.~Wang, Z.~Zhu, G.~Huang, X.~Chen, J.~Zhu, and J.~Lu.
\newblock Drivedreamer: Towards real-world-drive world models for autonomous driving.
\newblock In {\em European conference on computer vision}, pp. 55--72. Springer, 2024.

\bibitem{wang2025most}
Y.~Wang, Z.~Leng, F.~W. Li, X.~Liang, et~al.
\newblock Most: Motion diffusion model for rare text via temporal clip banzhaf interaction.
\newblock {\em IEEE Transactions on Visualization and Computer Graphics}, 2025.

\bibitem{wang2023fg}
Y.~Wang, Z.~Leng, F.~W. Li, S.-C. Wu, and X.~Liang.
\newblock Fg-t2m: Fine-grained text-driven human motion generation via diffusion model.
\newblock In {\em Proceedings of the IEEE/CVF International Conference on Computer Vision}, pp. 22035--22044, 2023.

\bibitem{wang2025fg}
Y.~Wang, M.~Li, J.~Liu, Z.~Leng, F.~W. Li, Z.~Zhang, and X.~Liang.
\newblock Fg-t2m++: Llms-augmented fine-grained text driven human motion generation.
\newblock {\em International Journal of Computer Vision}, pp. 1--17, 2025.

\bibitem{wang2024move}
Z.~Wang, Y.~Chen, B.~Jia, P.~Li, J.~Zhang, J.~Zhang, T.~Liu, Y.~Zhu, W.~Liang, and S.~Huang.
\newblock Move as you say interact as you can: Language-guided human motion generation with scene affordance.
\newblock In {\em Proceedings of the IEEE/CVF Conference on Computer Vision and Pattern Recognition}, pp. 433--444, 2024.

\bibitem{wang2022humanise}
Z.~Wang, Y.~Chen, T.~Liu, Y.~Zhu, W.~Liang, and S.~Huang.
\newblock Humanise: Language-conditioned human motion generation in 3d scenes.
\newblock {\em Advances in Neural Information Processing Systems}, 35:14959--14971, 2022.

\bibitem{wang2025spatial}
Z.~Wang, J.~Zhang, Y.~Chen, B.~Jia, W.~Liang, and S.~Huang.
\newblock Spatial-temporal multi-scale quantization for flexible motion generation.
\newblock {\em arXiv preprint arXiv:2508.08991}, 2025.

\bibitem{xing2024scene}
C.~Xing, W.~Mao, and M.~Liu.
\newblock Scene-aware human motion forecasting via mutual distance prediction.
\newblock In {\em European Conference on Computer Vision}, pp. 128--144. Springer, 2024.

\bibitem{zhang2023generating}
J.~Zhang, Y.~Zhang, X.~Cun, Y.~Zhang, H.~Zhao, H.~Lu, X.~Shen, and Y.~Shan.
\newblock Generating human motion from textual descriptions with discrete representations.
\newblock In {\em Proceedings of the IEEE/CVF Conference on Computer Vision and Pattern Recognition}, pp. 14730--14740, 2023.

\bibitem{zhang2022motiondiffuse}
M.~Zhang, Z.~Cai, L.~Pan, F.~Hong, X.~Guo, L.~Yang, and Z.~Liu.
\newblock Motiondiffuse: Text-driven human motion generation with diffusion model.
\newblock {\em IEEE Transactions on Pattern Analysis and Machine Intelligence}, pp. 1--15, 2024. \href{https://doi.org/10.1109/TPAMI.2024.3355414}
{doi: {{%
10\hspace{.1pt}\discretionary{.}{%
}{.}\hspace{.4pt}1109\discretionary{/}{%
}{/}TPAMI\hspace{.1pt}\discretionary{.}{%
}{.}\hspace{.4pt}2024\hspace{.1pt}\discretionary{.}{%
}{.}\hspace{.4pt}3355414}}}


\bibitem{zhang2023remodiffuse}
M.~Zhang, X.~Guo, L.~Pan, Z.~Cai, F.~Hong, H.~Li, L.~Yang, and Z.~Liu.
\newblock Remodiffuse: Retrieval-augmented motion diffusion model.
\newblock In {\em Proceedings of the IEEE/CVF International Conference on Computer Vision}, pp. 364--373, 2023.

\bibitem{zhang2024large}
M.~Zhang, D.~Jin, C.~Gu, F.~Hong, Z.~Cai, J.~Huang, C.~Zhang, X.~Guo, L.~Yang, Y.~He, et~al.
\newblock Large motion model for unified multi-modal motion generation.
\newblock In {\em European Conference on Computer Vision}, pp. 397--421. Springer, 2024.

\bibitem{zhang2020place}
S.~Zhang, Y.~Zhang, Q.~Ma, M.~J. Black, and S.~Tang.
\newblock Place: Proximity learning of articulation and contact in 3d environments.
\newblock In {\em 2020 International Conference on 3D Vision (3DV)}, pp. 642--651. IEEE, 2020.

\bibitem{zhao2022compositional}
K.~Zhao, S.~Wang, Y.~Zhang, T.~Beeler, and S.~Tang.
\newblock Compositional human-scene interaction synthesis with semantic control.
\newblock In {\em European Conference on Computer Vision}, pp. 311--327. Springer, 2022.

\bibitem{zhong2023attt2m}
C.~Zhong, L.~Hu, Z.~Zhang, and S.~Xia.
\newblock Attt2m: Text-driven human motion generation with multi-perspective attention mechanism.
\newblock In {\em Proceedings of the IEEE/CVF International Conference on Computer Vision}, pp. 509--519, 2023.

\bibitem{zhou2024golden}
Z.~Zhou, S.~Shao, L.~Bai, S.~Zhang, Z.~Xu, B.~Han, and Z.~Xie.
\newblock Golden noise for diffusion models: A learning framework.
\newblock {\em arXiv preprint arXiv:2411.09502}, 2024.

\end{thebibliography}

\end{document}